\documentclass{article}
\usepackage[final,dblblindworkshop]{neurips_2025}
\workshoptitle{Grounded and Faithful Vision-Language Models for Real-World Deployment}
\makeatletter
\renewcommand{\@neuripsordinal}{40th}
\renewcommand{\@neuripsyear}{2026}
\makeatother

\makeatletter
\renewcommand{\@neuripsordinal}{40th}
\renewcommand{\@neuripsyear}{2026}
\makeatother

\usepackage[T1]{fontenc}
\usepackage{amsmath}
\usepackage{booktabs}
\usepackage{url}
\usepackage{amssymb}
\usepackage{amsthm}
\usepackage{graphicx}
\usepackage{multirow}
\usepackage{enumitem}
\usepackage{xcolor}
\usepackage{cleveref}

\theoremstyle{plain}

\theoremstyle{definition}

\crefname{section}{Sec.}{Secs.}
\crefname{table}{Tab.}{Tabs.}
\crefname{figure}{Fig.}{Figs.}
\crefname{equation}{Eq.}{Eqs.}

\newcommand{\pbar}{\bar p}

\newcommand{\E}{\mathbb{E}}

\DeclareMathOperator*{\argtopm}{arg\,top\text{-}\emph{m}}

\title{What Entropy-Based View Selection Cannot See:\\
       An Audit of Test-Time Adaptation for Vision-Language Models}

\author{%
  Sivasuthan Sukumar \\
  University of Moratuwa \\
  Sri Lanka \\
  \texttt{sivasuthan.26@cse.mrt.ac.lk}
  \And
  Uthayasanker Thayasivam \\
  University of Moratuwa \\
  Sri Lanka \\
  \texttt{rtuthaya@cse.mrt.ac.lk}
}

\begin{document}
\maketitle

\begin{abstract}
A vision-language model such as CLIP can classify images into categories it was never explicitly trained to recognize, but accuracy degrades when test images differ in style or source from the training data. A widely used remedy generates randomly cropped and flipped copies of each test image, called views, and combines their predictions. Because some views miss the object, these methods must decide which views to trust. The dominant approach measures prediction entropy, the concentration of each view's predicted class distribution, and retains the most certain views. Other methods use quantities such as how typical a view's embedding is among views of the same image. Despite repeated refinement of these scoring functions, the information they use, the model's own response to each view, has remained largely unexamined. We examine this choice. Holding each stage of the pipeline fixed, we first show that view selection, rather than the gradient update, produces almost all of the accuracy gains. We then ask whether entropy can identify useful views when the method is wrong despite at least one view predicting the correct class. On these cases, entropy ranks correct views above incorrect ones no better than chance, recovering only about one-twentieth of the accuracy recovered by an oracle over the same views. Entropy is therefore weakly informative rather than useless. This result holds for two image encoders and under a control ruling out an obvious statistical artifact. An exact analysis of the entropy objective explains why: the score measures confidence, but not which class a view favors. We identify an unused signal in the pipeline: the augmentation procedure already knows how much of the image each crop contains. Ranking views by this quantity improves accuracy on 14 of 15 benchmark datasets over entropy and on 11 of 15 over the strongest scoring function tested. A pre-method measurement also predicts which datasets benefit.

\end{abstract}

\section{Introduction}
\label{sec:intro}

Vision-language models such as CLIP \citep{radford2021clip} learn a shared embedding space for
images and text, which lets them classify an image into any set of categories named in words,
without task-specific training. Their accuracy drops when test images differ in style, rendering
or source from the training data, a \emph{distribution shift}. Since a zero-shot model is
deployed precisely because the target distribution is unknown in advance, this is the operating
regime rather than an edge case.

\begin{figure}[t]
  \centering
  \includegraphics[width=\linewidth]{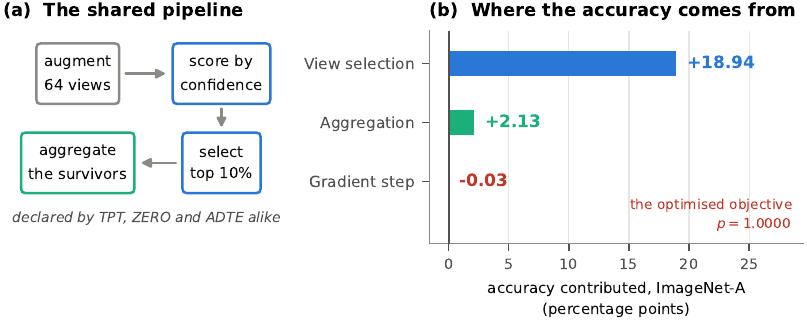}
  \caption{\textbf{The pipeline, and which of its stages produces the accuracy.} \textbf{(a)} The
  augment, score, select, aggregate recipe TPT, ZERO and ADTE each declare. \textbf{(b)} Each
  stage measured with the other two held fixed, on ImageNet-A, in \emph{percentage points}: the
  absolute gap between two accuracies, so $40\%$ to $59\%$ is $19$ of them. Selection produces
  almost all of the gain; the gradient step is indistinguishable from no update under a paired
  McNemar test \citep{mcnemar1947}. Concurrent work agrees by different means
  \citep{huang2026whatdrives}; we claim the decomposition and its mechanism, not the direction.}
  \label{fig:pipeline}
\end{figure}

\paragraph{The pipeline this paper studies.}
One family of remedies, \emph{test-time adaptation}, adjusts the prediction for each test image
separately, using no labels. Within it we study the variant built on augmentation
(\cref{fig:pipeline}a): random crops and horizontal flips turn the single test image into $N$
transformed copies, called \emph{views}. Some views contain the object clearly, others crop it
away entirely, so the method scores each view and keeps only the highest-scoring ones. The
dominant line of methods computes that score from one quantity, the \emph{prediction entropy}
$H(p_i) = -\sum_c p_{ic}\log p_{ic}$, where $p_i$ is the model's predicted distribution over
classes for view $i$. Entropy is near zero when almost all probability sits on one class and
large when it is spread across many, so a low-entropy view is one the model is confident about.
The surviving views are averaged into a single distribution, and some methods additionally take
one gradient step lowering the entropy of that average. A smaller line replaces entropy with a
different model-derived quantity, for instance the density of a view's embedding among the
embeddings of the other views \citep{zanella2024mta}. Our audit targets the \emph{input} these
scores share, the model's own response to the view, so it bears on both lines;
\cref{sec:related} separates what we test from what we position against.

The scope of this pipeline is not something we assert; the methods state it, quoting their own
setup sections (\cref{app:repro}). ZERO reports that it
\emph{``inherit[s] the setup of TPT with $N=64$''} \citep{farina2024zero}, and ADTE adopts that
same setup with a filtering ratio of $0.1$ \citep{wu2026adte}. TPT \citep{shu2022tpt} originated
the recipe and later work inherits it (\cref{sec:related}). Batch-level adaptation
\citep{wang2021tent} and cache-based methods \citep{karmanov2024tda} lie outside this scope and
we claim nothing about them.

\paragraph{Which stage matters, and how we isolated it.}
Before asking whether the view score is good, we established that the score is what governs the
pipeline. The three stages vary independently, so we measured each while holding the other two
fixed and reusing identical views, cached embeddings and readout, which attributes any difference
to the stage varied rather than to the data or the evaluation. Selection accounts for $18.94$
percentage points, aggregation for $2.13$, and the gradient step for $-0.03$, indistinguishable
from making no update (\cref{fig:pipeline}b). The score, not the optimisation, is what these
methods run on.

\paragraph{What subsequent research changed, and what it did not.}
Later studies improved the scoring \emph{function}: a zero-temperature average
\citep{farina2024zero}, Tsallis entropy \citep{tsallis1988}, a one-parameter generalisation of
Shannon entropy, and a class-adaptive Tsallis exponent \citep{wu2026adte}. Each keeps the
scoring \emph{input} fixed at the model's own output distribution on the view. Whether that
input carries the information view selection needs has not been tested, and testing it is the
subject of this paper.

\begin{figure}[t]
  \centering
  \includegraphics[width=\linewidth]{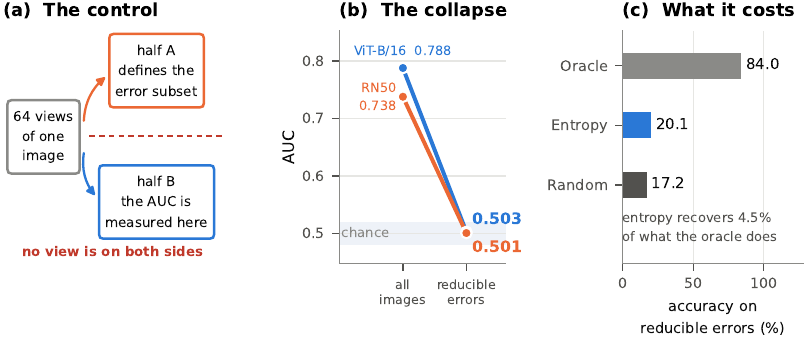}
  \caption{\textbf{The entropy score separates useful views from useless ones only where a better
  choice cannot help.} \textbf{(a)} The control: reducible errors are defined on one random half
  of an image's views and the score evaluated on the disjoint half, so restricting the population
  cannot contaminate the measurement. \textbf{(b)} Area under the ROC curve (AUC,
  \cref{sec:audit}), where $1.0$ orders correct views perfectly above incorrect ones and $0.5$
  carries no information; on reducible errors it falls to chance on both encoders.
  \textbf{(c)} The consequence in accuracy on those images.}
  \label{fig:collapse}
\end{figure}

\paragraph{Testing the input, and the control the test requires.}
A better view score can only help on images where a better choice of views exists. We call an
error \emph{reducible} if the averaged prediction is wrong while at least one of the $N$ views
already predicts the correct class; $49$ to $73\%$ of all errors are of this kind. Across all
test images the entropy score ranks correct views above incorrect ones well; on reducible errors
it is at chance (\cref{fig:collapse}b), under the control \cref{sec:audit} builds to exclude the
one objection that comparison must survive. The score is informative where selection cannot
change the answer, and uninformative where it could.

\paragraph{A signal of a different kind, already present.}
The augmentation procedure \emph{sampled} the crop box for each view, so the pipeline already
records how much of the image each view contains. That measurement needs no labels, costs nothing
to store and, unlike every score above, is not computed from the model's response; of the
test-time adaptation methods for vision-language models we surveyed, none reads it back. Ranking views with it raises accuracy on the standard $15$-dataset
benchmark, and a label-free measurement predicts in advance which datasets benefit.

\paragraph{Contributions.}
\textbf{(i)} A controlled audit of the view score under the conditioning that makes it
meaningful, showing it is at chance where the remaining accuracy lies, on two image encoders
(\cref{sec:audit}). \textbf{(ii)} An exact analysis of what a score computed from the model's
output can express, valid for any tuned parameters, with a matched comparison isolating the
missing ingredient (\cref{sec:theory,sec:audit}). \textbf{(iii)} Selection from augmentation
parameters as a proof of concept, with per-dataset gains predictable in advance and the regime
in which it does not help (\cref{sec:method,sec:exp,sec:disc}).

\section{Problem formulation}
\label{sec:setup}

\paragraph{The failure mode in one sentence.}
Because augmentation makes CLIP overconfident rather than merely noisy
\citep{farina2024zero}, a view covering $8\%$ of the image and landing on background can receive
a confident prediction and be kept over a view containing the object, even though the pipeline
recorded that it saw $8\%$ of the image when it drew the box.

\paragraph{Notation.}
Let $v_i$ be the image embedding of view $i$ and $w_c(\theta)$ the text embedding of class $c$,
from a prompt with tunable parameters $\theta$. The class scores, or \emph{logits}, are
$z_{ic}=s\langle v_i,w_c(\theta)\rangle$ with $s$ a fixed temperature, and
$p_i=\mathrm{softmax}(z_i)$ is view $i$'s distribution over the $K$ classes. Writing $S$ for the
selected set, $\pbar=\frac{1}{|S|}\sum_{i\in S}p_i$ is the averaged prediction the method answers
with. The image encoder runs without gradient tracking, so every $v_i$ is constant in $\theta$
and the analysis of \cref{sec:theory} is exact rather than approximate. Methods that tune
parameters minimise $L=H(\pbar)$; ZERO and ADTE tune nothing and select only.

\paragraph{Two selection rules.}
The rule used throughout this literature ranks views by a function of $p_i$ alone and keeps the
$m$ best,
\begin{equation}
  S \;=\; \argtopm_i \; \big[-H(p_i)\big].
  \label{eq:entsel}
\end{equation}
The crop box that produced view~$i$, giving its position and size in normalised image
coordinates, costs nothing to record and is statistically independent of how the model responds.
We combine the two signals by converting each to a within-image rank first, since an entropy
value and an area are in different units and cannot be added directly:
\begin{equation}
  S \;=\; \argtopm_i \;
  \big[(1-\omega)\,r_i^{\mathrm{ent}} + \omega\, r_i^{\mathrm{area}}\big],
  \label{eq:ours}
\end{equation}
where $r^{\mathrm{ent}}_i$ is the rank of $-H(p_i)$ among the views of that image,
$r^{\mathrm{area}}_i$ the rank of the crop area $h_i w_i$, and $\omega \in [0,1]$ the mixing
weight. Setting $\omega=0$ recovers \cref{eq:entsel} exactly, so the standard rule is a special
case of ours.

\section{Auditing the view score}
\label{sec:audit}

\subsection{How much accuracy a better choice of views could recover}

Selection can only help on a reducible error, so we first measure what those are worth. Such an
error needs no better model and no better augmentation, only a better choice among views already
computed. To bound what that choice could buy we use an \emph{oracle} selector, which keeps the
same number of views but picks them using the ground-truth label. An oracle is not a method,
since it needs labels a deployed system lacks; it is an upper bound on any rule operating on the
same views. The gap to it runs from $12.4$ to $26.8$ percentage points, and $49$ to $73\%$ of
errors are reducible (\cref{app:ceiling}). The gap is \emph{larger} on the weaker encoder, which
rules out reading it as an artefact of an already-strong model.

\subsection{Can the entropy score reach that headroom?}

\paragraph{The quantity we measure.}
Each view produces its own prediction and is individually right or wrong, and any score induces a
ranking over the views of an image. The natural measure of whether that ranking is useful is the
\emph{area under the ROC curve}, written AUC: the probability that a randomly chosen correct view
is scored better than a randomly chosen incorrect one, so a perfect ranking gives $1.0$ and an
uninformative one $0.5$. It is computed per image with no method in the loop, so it evaluates the
score itself rather than a pipeline built on it.

\paragraph{The objection this measurement must survive, and the control it forces.}
Reducible errors are defined using the averaged prediction $\pbar$, and confident views agree
with one another, so conditioning on ``the average is wrong'' could select images on which
confident views are wrong together, manufacturing a low AUC even if confidence and correctness
were positively related within every image. That entropy never references $\pbar$ does not settle
it, because the \emph{subset} does, and only a design that breaks a statistical dependence
removes it. We therefore split the $64$ views of each image at random into halves $A$ and $B$:
half $A$ alone decides whether the image is a reducible error and the AUC is computed only on
half $B$, so if the confounder were producing the effect the controlled AUC would be
substantially higher than the uncontrolled one.

\paragraph{Result.}
It is not (\cref{fig:collapse}b; full table in \cref{app:ceiling}). The controlled AUC is $0.503$ on the vision
transformer and $0.501$ on the convolutional encoder, against $0.788$ and $0.738$ across all
images; controlled and uncontrolled values agree within $0.003$ on all twelve runs, so the
confounder is not the explanation, and two encoders with different inductive biases reach chance
independently. The failure is, however, specific to the \emph{output distribution}. Statistics
from the cached image embeddings, such as how central a view's embedding is and how similar it is
to the unmodified image, score $0.454$ and $0.468$ while correlating only weakly with entropy,
Spearman $\rho$ between $0.19$ and $0.45$. Both sit \emph{below} chance on both encoders, so
their ordering is mildly anti-predictive and reversing it gives $0.546$ and $0.532$, a larger
displacement from chance than any score computed from the posterior. We built no selector on the reversed statistic and report it as an observation, not a result. It
indicates that usable signal survives outside the model's response, the hypothesis
\cref{sec:method} tests.

\paragraph{What this licenses, and what it does not.}
An AUC at chance for \emph{ranking} does not by itself mean entropy selection is no better than
random selection, and we tested the difference. Under the same split-half design entropy
selection reaches $20.15\%$ against $17.17\%$ for random, an advantage of $2.98$ percentage
points with a $95\%$ bootstrap interval of $[1.28, 4.72]$, while an oracle over the same views
reaches $84.03\%$: the score captures about $4.5\%$ of the accuracy a better choice could recover
(\cref{fig:collapse}c). The claim is quantitative, that the score is weakly informative here
rather than useless.

\subsection{What the model's output omits}

\Cref{sec:theory} shows that the standard view scores, including Shannon entropy, Tsallis and
R\'enyi entropy at fixed order, maximum probability and the top-two margin, are unchanged if the
class labels are permuted. Such a score is therefore a function of the \emph{sorted} probability
vector alone: it can report how concentrated a view is, but not which class the view favours.

To measure what that restriction costs we compare two arms differing in exactly one respect:
both fit the same regressor on the same rows, from feature vectors of \emph{identical dimension
containing the identical multiset of numbers}, sorted by magnitude in one arm and indexed by
class in the other, so a gap cannot come from dimensionality, capacity or the values themselves
(\cref{app:matched2}). Preserving class identity is worth $8.60$ percentage points, positive in
$\mathbf{40}$ of $\mathbf{40}$ configurations across $8$ datasets and $5$ encoders, with a
capacity ladder saturated throughout (\cref{app:matched}). This bounds the \emph{information}
available, not a deployable method, since both arms fit with labels; ADTE's class-specific
exponent is the one published score breaking the symmetry, and it buys roughly $0.5$ percentage
points end to end \citep{wu2026adte}.

The same measurement accounts for findings we had obtained separately and could not connect,
including why no selector we constructed improved on entropy, and it extends to a different
parameter group: MEMO's gradient step \citep{zhang2022memo} improves the averaged prediction in
$0$ of $54$ configurations (\cref{app:explains}).

\section{What a score computed from the model's output can express}
\label{sec:theory}

The audit in \cref{sec:audit} is an observation. This section explains it and states how far the
explanation reaches. Full derivations are in \cref{app:proofs}; each result names the subsection
that proves it.

\paragraph{Why we analyse the score in logit space, and what that buys.}
Methods in this family tune different parameters, so an analysis tied to one parameter set would
not transfer. We avoid that by differentiating with respect to the \emph{logits} $z$ rather than
the parameters $\theta$. For any differentiable model the chain rule gives
$\nabla_\theta L = J^\top \nabla_z L$ with $J = \partial z/\partial\theta$, so $\nabla_z L$
is a factor of \emph{every} parameter-space update, whichever parameters are tuned, and
statements about it hold for any gradient-based minimiser of $H(\pbar)$ (\cref{app:reach}). This
is the factorisation on which kernel analyses of training dynamics are built
\citep{jacot2018ntk}; we use it and not the infinite-width theorem, whose assumptions we
measured and found violated here. \Cref{app:whylogit} gives that argument in full with the line
it forbids us to cross: our statements are about the force $\partial L/\partial z$, not about
which logits actually rise. Because the image embeddings are constants, the views reach $\theta$
only through one matrix $\Psi$ (\cref{app:psi}). Every identity below was checked against
automatic differentiation in double precision (\cref{tab:resid}).

\paragraph{The gradient transmits a ranking, not a margin.}
Differentiating $L=H(\pbar)$ through the softmax gives, for view $j$ and class $c$,
\begin{equation}
  \frac{\partial L}{\partial z_{jc}}
  = -\frac{1}{N}\sum_{c'} p_{jc}\,p_{jc'}\,\log\frac{\pbar_c}{\pbar_{c'}} .
  \label{eq:thmA}
\end{equation}
The derivation is in \cref{app:gradient}. Read the right-hand side as a sum over pairs of
classes, each contributing a force with two separable parts. The factor $p_{jc}p_{jc'}$ is a
gate, negligible unless view $j$ assigns real probability to \emph{both} classes, so only pairs
the view is undecided about contribute. The factor $\log(\pbar_c/\pbar_{c'})$ supplies
direction and size, and its sign is simply whether $\pbar_c$ exceeds $\pbar_{c'}$.

That second part is the important one, and it must be stated precisely. The \emph{sign} of each
pairwise term is $\mathrm{sign}(\pbar_c-\pbar_{c'})$, a rank statistic of $\pbar$ that records
which class the average prefers and nothing about how reliable that preference is; the
\emph{magnitude} is not a rank statistic, since $|\log(\pbar_c/\pbar_{c'})|$ grows as the
average sharpens. A near-tie therefore hands the update a direction fixed by an arbitrarily small
margin, damped only logarithmically. We claim this of the pairwise terms, not of the assembled
gradient vector, which sums magnitudes over all pairs before the optimiser acts on it. A view
whose distribution is exactly one-hot
contributes nothing at all, since $p_{jc}p_{jc'}=0$ for every pair, and because $J^\top 0 = 0$
for any Jacobian that holds whatever parameters are tuned. Keeping only the most confident views
therefore keeps precisely the views that contribute least to the update.

\paragraph{The objective does not enforce agreement between views.}
It is often assumed that minimising the entropy of the average makes the views agree. It does not
follow. Let $b$ be a class view $j$ favours but the average does not, write
$L_c=\log(\pbar_c/\pbar_b)$, let $\tilde p_j$ be $p_j$ conditioned off $b$, and let
$M_j(s)=\E_{c\sim\tilde p_j}[e^{sL_c}]$ be the moment generating function of those log-ratios.
Let $\Phi_{jb} := -\partial L/\partial z_{jb}$ denote the \emph{force} on class $b$, the sign
convention used throughout. In Tsallis form of order $q$, which recovers Shannon entropy as
$q\to1$ \citep{tsallis1988}, the update drives view $j$ \emph{further} towards its dissenting
class exactly when
\begin{equation}
  \Phi_{jb} > 0 \iff
  \big[\,\E_{\tilde p_j}[L] < 0 \ \ (q{=}1);\quad
        M_j(q-1) < 1 \ \ (q{>}1);\quad
        M_j(q-1) > 1 \ \ (q{<}1)\,\big].
  \label{eq:thmB}
\end{equation}
The cases do not collapse: $M_j(0)=1$ holds identically for any moment generating function, so
the $q>1$ inequality degenerates exactly at the Shannon order every method in scope uses, leaving
the mean log-ratio as the operative condition there. $M_j$ is strictly convex with
$M_j'(0)=\E_{\tilde p_j}[L]$, so when the Shannon case fires there is a unique positive root
$s^\star$ and a critical order $q^\star=1+s^\star$, classically the Lundberg coefficient
\citep{asmussen2010ruin}. The reversal below $q=1$ is why an order good for \emph{scoring} views
need not be good for \emph{driving} an update (\cref{sec:exp}). \Cref{app:cramer} proves all
three cases and exhibits a step that lowers the loss while moving a view further from the
average, so agreement is not something the objective enforces. Verified on $279$ dissenting
views at six Tsallis orders, $1674$ checks, with no sign disagreements.

\paragraph{Little of the gradient survives to the parameters.}
Much of \cref{eq:thmA} is discarded before reaching the parameters: the optimiser is reset per
image, so each update is a first step from zero moments, which for Adam is the \emph{sign} of the
gradient \citep{balles2018adam}, and what survives passes through $\Psi$, whose singular values
are dominated by one direction. Unlike the results above this depends on the optimiser and the
parameter group, and realised logit motion agrees with the force coordinatewise only part of the
time, so we give it as measurement, not theorem (\cref{app:optimiser}).

\paragraph{Together.}
A signal whose direction encodes only the average's ranking, whose magnitude the optimiser
discards, and which reaches the image through an essentially one-dimensional channel cannot
repair a badly chosen set of views. This supplies the mechanism behind the reported
near-invariance of the averaged prediction to the update \citep{farina2024zero}.

\section{Selecting views from the augmentation parameters}
\label{sec:method}

Two quantities in that audit are displaced from chance (\cref{tab:blind}): crop area, at
$0.510$ against $0.503$ for entropy, and the embedding statistics, which are anti-predictive and
would have to be reversed. Crop area is the only one that is usable as scored \emph{and} not
computed from the model's response, so it is the one we test. That test is direct: if what the
entropy score lacks is information the model's output cannot carry, the simplest available
quantity of a different kind should help. What follows is that test, not a tuned system.

\paragraph{The signal.}
The augmentation operator samples a crop box with area fraction drawn uniformly from
$[0.08,1.0]$ and resizes to $224\times224$ pixels. The pipeline generates this box, so recording
its area costs nothing and needs no labels. We rank views by area and combine that
ranking with the entropy ranking as in \cref{eq:ours}. The mixing weight $\omega$ and the number
of retained views $m$ are chosen by $5$-fold cross-fitting: the parameters scoring each fifth of
the data are chosen using only the other four fifths, so no image is scored under parameters
selected with its own outcome. The baseline is the standard entropy rule at its published
setting and is never tuned. We state that asymmetry plainly rather than claim it favours the
baseline: our rule fits two parameters where the baseline fits none, so part of any gain could
be the freedom to choose $m$. What bounds the concern is that $\omega=0$ recovers the baseline
exactly, and that the per-dataset gains are forecast by an AUC measured with no method in the
loop, which a free parameter cannot produce.

\paragraph{Predicting in advance where the signal exists.}
A method that helps on some datasets and hurts on others is useful only if one can tell which
beforehand. The controlled AUC of \cref{sec:audit} does that with no method in the loop: across the $15$
datasets the crop-area AUC on reducible errors predicts the eventual gain with a Spearman rank
correlation of $\rho = 0.949$, where $\rho$ measures how well one ordering matches another on a
scale from $-1$ to $1$ (\cref{fig:law}). Datasets whose
crop-area AUC falls below chance gain nothing, and cross-fitting drives $\omega$ towards zero
on them. That protection is not exact. DTD, the single dataset that loses, is one of them, and
it loses $0.8$ percentage points; with no signal for $\omega$ to exploit, that loss is
attributable to the cross-fitted $m$ rather than to crop area, and we count it against the
method rather than explaining it away.

This is what separates the result from a tuning artefact: seven significant gains among fifteen
datasets could arise from testing many datasets, but an independent measurement that forecasts
the \emph{size} of each gain could not. It also survives the circularity objection that the AUC
and the gain use the same images, since computing them on \emph{disjoint} halves still leaves
$\rho = 0.881$.

\paragraph{What we do not claim.}
We have no mechanism: two accounts written down before testing were both rejected
(\cref{app:failed}). We report the empirical relationship, which is strong and predictive, and
label the mechanism unknown rather than offer a third untested account.

\section{Experiments}
\label{sec:exp}

\begin{figure}[t]
  \centering
  \includegraphics[width=0.72\linewidth]{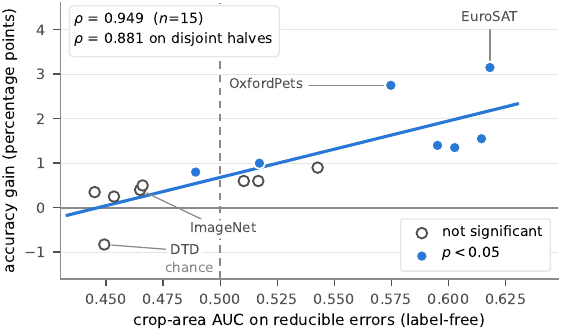}
  \caption{\textbf{The gain is predictable before the method is run.} One point per dataset; the
  horizontal axis is how well a view's \emph{crop area} separates correct from incorrect views on
  reducible errors, an AUC computed with no method in the loop. Datasets left of the chance line
  gain nothing, as a real signal rather than a tuning artefact should. Per-dataset gains are in
  \cref{tab:main}.}
  \label{fig:law}
\end{figure}

\paragraph{Protocol, and what it holds fixed.}
We use the $15$-dataset benchmark standard in this literature, ImageNet
\citep{deng2009imagenet} with four distribution shifts plus ten fine-grained datasets, listed in
\cref{app:repro}. Backbones are CLIP ViT-B/16 and RN50 \citep{radford2021clip}; the
matched-information control additionally spans MetaCLIP, SigLIP and ViT-L/14. Every arm sees \emph{identical views, the same cached image embeddings and the same
readout}, so an accuracy difference can only come from the selection rule. Significance uses a
paired McNemar test \citep{mcnemar1947}, which counts only the images on which two classifiers
disagree; AUC intervals are bootstrap estimates over $2000$ resamples; selector parameters are
cross-fitted and baselines are never tuned. All $15$ loaders were verified against published
zero-shot accuracy at full test-set size, matching on ImageNet to within $0.01$ percentage
points. The comparisons replay a cached tensor of $2000$ images per dataset, drawn once and stratified
by class on ImageNet, with \emph{every arm scored on identical images}, so the sample cannot
favour a rule; it does bound resolution to about half a percentage point under a paired test. We
read smaller per-dataset differences as directional and rest the claim on the suite average, the
sign count and the forecast of \cref{fig:law}. \Cref{app:design} states, for each measurement and each analytical
assumption, the question it answers, the alternative explanation it excludes, and why its metric
is the right one.

\begin{table}[t]
  \centering\scriptsize
  \setlength{\tabcolsep}{1.15pt}
  \caption{\textbf{Selection rules on the $15$-dataset benchmark}, CLIP ViT-B/16, with identical
  views, cache and readout so that \emph{only} the selection rule differs. The baseline is never
  tuned; our parameters are cross-fitted. $^*$ marks $p<0.05$ under paired McNemar
  \citep{mcnemar1947}. Crop geometry is positive on $14$ of $15$ datasets and its largest gains
  are those the label-free measurement flags in advance (\cref{fig:law}).}
  \label{tab:main}
  \vspace{2pt}
  \begin{tabular}{@{}lcccccc@{\hskip 4pt}cccccccccc@{\hskip 3pt}c@{}}
    \toprule
    & \multicolumn{6}{c}{\textbf{ImageNet and distribution shifts}}
    & \multicolumn{10}{c}{\textbf{Fine-grained}} & \\
    \cmidrule(lr){2-7}\cmidrule(lr){8-17}
    Selection rule & IN & IN-A & IN-K & IN-V2 & IN-R & \textit{Avg.}
    & Flwr & DTD & Pets & Cars & UCF & Cal & Food & SUN & Air & ESAT & \textbf{All} \\
    \midrule
    Shannon entropy \citep{shu2022tpt}
      & 68.5 & 59.6 & 46.4 & 61.5 & 77.3 & \textit{62.7}
      & 67.0 & 45.9 & 86.0 & 66.8 & 67.0 & 92.6 & 84.7 & 60.9 & 24.4 & 39.4 & 63.2 \\
    Tsallis $q{=}0.5$
      & \textbf{68.9} & \textbf{60.8} & 47.0 & 61.6 & 77.8 & \textit{63.2}
      & 66.8 & \textbf{46.2} & 87.0 & 67.3 & 67.0 & 92.7 & 84.5 & \textbf{61.4} & 24.2 & 40.2 & 63.6 \\
    \textbf{Crop geometry (ours)}
      & \textbf{68.9} & 59.9 & \textbf{47.8}$^*$ & \textbf{62.0} & \textbf{78.7}$^*$ & \textit{\textbf{63.4}}
      & \textbf{68.0}$^*$ & 45.1 & \textbf{88.8}$^*$ & \textbf{67.8} & \textbf{67.6} & \textbf{94.2}$^*$
      & \textbf{85.5}$^*$ & 61.2 & \textbf{25.1} & \textbf{42.5}$^*$ & \textbf{64.2} \\
    \bottomrule
  \end{tabular}
\end{table}

\paragraph{Main result.}
\Cref{tab:main} compares selection rules with everything else held fixed, so the differences
isolate the rule. Against the standard Shannon rule, ranking with crop area raises the suite
average by $0.98$ percentage points, improves $14$ of $15$ datasets, is significant on $7$ and
significantly worse on none. Against the stronger Tsallis score the margin is smaller and we
state it in full: $11$ wins, one tie and three losses, averaging $0.6$ percentage points. The
asterisks are paired tests against Shannon; we ran none between the two non-baseline rows, so
that head-to-head is counts and a margin, not a significance claim. The point is not that crop
area is the better score, but that a quantity carrying no information about the model's response
competes with the best functional of that response.

\paragraph{Does the effect depend on the image encoder?}
Repeating the evaluation on the convolutional encoder RN50 gives, on the six datasets run with
both encoders, a mean gain of $1.18$ percentage points against $0.82$ for the vision transformer
on those same six, so the effect is not tied to one architecture and
is if anything stronger on the weaker encoder, consistent with its larger headroom, and the
predictive relationship holds within each encoder separately. The per-dataset gains themselves do
\emph{not} transfer: their rank correlation across encoders is $0.314$, so a dataset's gain is
not a constant of it and we quote none (\cref{app:backbones}).

\paragraph{A better scoring function is not a substitute.}
Improving the scoring \emph{function} while keeping its input fixed does help: a Tsallis score
at fixed order $q=0.5$ beats Shannon entropy on $11$ of $15$ datasets, ties on one and loses on
three (\cref{tab:main}), in the direction of ADTE's own ablation \citep{wu2026adte}. That order is harmful when the same functional drives an
optimisation step (\cref{sec:theory}), so the scoring and driving roles do not share an optimum.
Changing the input remains the larger effect; \cref{sec:disc} reports what happens when the two
are combined.

\section{Related work}
\label{sec:related}

We organise by \emph{what a method scores views on}, the axis this paper is about;
\cref{app:related} gives the extended discussion.

\paragraph{Scoring from the model's output: the incumbent.}
Entropy minimisation on unlabelled data originates in semi-supervised learning
\citep{grandvalet2004entropy} and became the standard test-time objective through TENT
\citep{wang2021tent} and SHOT \citep{liang2020shot}. MEMO \citep{zhang2022memo} moved it to the
single-image setting and TPT \citep{shu2022tpt} specialised it to vision-language models,
adapting the prompt tokens of \citet{zhou2022coop}. The scoring \emph{function} has since been
improved repeatedly, by a zero-temperature average \citep{farina2024zero}, Tsallis entropy
\citep{tsallis1988} and a class-adaptive Tsallis exponent \citep{wu2026adte}, while other work
varies the objective, the calibration or the views \citep{yoon2024ctpt,feng2023difftpt}. All keep
the scoring \emph{input} fixed at the model's output distribution, which is what our analysis
targets; the families excluded in \cref{sec:intro} are untouched by our claims
\citep{karmanov2024tda,zhu2024frolic}.

Calibration-oriented methods built on top of TPT, such as A-TPT \citep{ahamed2026tpt} and
AGI-TPT \citep{anonymous2026agitpt}, with related work on calibrated prompt tuning
\citep{ahamed2026m,sharifdeen2026mvc}, target calibration rather than view ranking and are
complementary to our audit (\cref{app:related}).

\paragraph{Scoring from the model's representation.}
A second line keeps the pipeline but replaces the output-distribution score: MTA
\citep{zanella2024mta} inherits the same $64$-view protocol and weights views by embedding
density, reporting that this beats confidence filtering; R-TPT \citep{sheng2025rtpt} reweights
views for robustness and reward-model feedback \citep{zhao2024rlcf} replaces entropy outright.
Not being functions of the sorted posterior these fall outside the class our symmetry argument
covers, and we do not claim it governs them. They are inside the audit: the embedding statistics
of \cref{tab:blind} are of this kind and are anti-predictive on reducible errors. Whether MTA's
particular score behaves the same way requires running it, and is the first experiment we would
add.

\paragraph{Questioning the update, and generation-side information.}
ZERO \citep{farina2024zero} proves the argmax of the averaged prediction is largely invariant to
the entropy update; we take that as established and do not re-derive it, adding the pairwise
decomposition of \cref{eq:thmA}, the dissent criterion of \cref{eq:thmB} and the stage-wise
attribution \citep[see also][]{huang2026whatdrives}. AugMix \citep{hendrycks2020augmix} and
\texttt{RandomResizedCrop} serve throughout this literature purely as generators; augmentation
identity has been read back when aggregating test-time predictions elsewhere in vision
\citep{shanmugam2021tta}, but within test-time adaptation for vision-language models, to our
knowledge, the sampled parameters are not. That the optimal error--reject rule thresholds the
true posterior is classical \citep{chow1970}; what it does not supply, and what we add, is the
protocol of conditioning on the subpopulation where the intervention can act.

\section{Discussion}
\label{sec:disc}

\paragraph{How the analysis and the measurements relate.}
We are explicit about which way the support runs. \Cref{eq:thmA} is exact and says a one-hot view
contributes zero, so a confidence filter retains the views contributing least; the measured
gradient stage is $-0.03$ percentage points. That measurement is the evidence and the analysis is
the account of it, not its derivation: the deployed step takes the \emph{sign} of the gradient,
and realised logit motion agrees with the force coordinatewise only $61.5\%$ of the time
(\cref{app:optimiser}), which is why we do not claim to have proved the update inert.
\Cref{eq:thmA,eq:thmB} concern the gradient and hold for any tuned parameters; the inertness is
an empirical finding on two parameter groups. Five hypotheses of our own were rejected by the
controls built to test them (\cref{app:failed}), including a stronger version of this paper's own
headline.

\paragraph{What this means for deployment.}
The backward pass can be deleted: it costs most of the compute and contributes $-0.03$
percentage points. Confidence remains a usable coarse filter but should not be trusted as a
per-view reliability signal on the samples that matter, where it recovers about one twentieth of
what a better choice among the same views is worth. And the augmenter records a label-free signal
no method we surveyed reads back. The failure is not in the model's perception but in the
\emph{reliability signal} built on it, invisible to any evaluation that does not condition on the
errors a better choice could still fix.

\paragraph{Limitations and scope.}
Crop geometry does not compose with a stronger scoring functional. We pre-registered the two
channels as additive; they are not, gaining $1.11$ percentage points against a prediction of
$1.81$, and under a stricter protocol the combination beats the better single signal in
$\mathbf{0}$ of $\mathbf{8}$ configurations, so once a stronger scoring function is in use the
crop-area gain largely does not add on top of it. We also have no mechanism for the gain itself.
Our rule cross-fits two parameters against an untuned baseline, so we do not claim it is
tuning-free, and we have not isolated the contribution of the retained-view count from that of
the crop-area weight; the DTD loss is the visible cost of that choice. Comparisons replay a
fixed cached subsample rather than the full test sets, which bounds the resolution of the paired
tests. The selection results run on the shared configuration TPT, ZERO and ADTE each declare,
but we did not run their code, so that reach is an argument rather than a measurement
(\cref{app:adte}). The gradient-stage results are specific to the optimiser and parameter group
studied, while \cref{eq:thmA,eq:thmB} hold for any tuned parameters.

{\small
\bibliographystyle{plainnat}
\bibliography{main}}

@inproceedings{radford2021clip,
  title     = {Learning Transferable Visual Models From Natural Language Supervision},
  author    = {Radford, Alec and Kim, Jong Wook and Hallacy, Chris and Ramesh, Aditya and
               Goh, Gabriel and Agarwal, Sandhini and Sastry, Girish and Askell, Amanda and
               Mishkin, Pamela and Clark, Jack and Krueger, Gretchen and Sutskever, Ilya},
  booktitle = {ICML},
  year      = {2021}
}

@inproceedings{shu2022tpt,
  title     = {Test-Time Prompt Tuning for Zero-Shot Generalization in Vision-Language Models},
  author    = {Shu, Manli and Nie, Weili and Huang, De-An and Yu, Zhiding and
               Goldstein, Tom and Anandkumar, Anima and Xiao, Chaowei},
  booktitle = {NeurIPS},
  year      = {2022},
  note      = {arXiv:2209.07511}
}

@inproceedings{farina2024zero,
  title     = {Frustratingly Easy Test-Time Adaptation of Vision-Language Models},
  author    = {Farina, Matteo and Franchi, Gianni and Iacca, Giovanni and
               Mancini, Massimiliano and Ricci, Elisa},
  booktitle = {NeurIPS},
  year      = {2024},
  note      = {arXiv:2405.18330}
}

@inproceedings{wu2026adte,
  title     = {Adaptive Debiasing Tsallis Entropy for Test-Time Adaptation},
  author    = {Wu, Xiangyu and Jiang, Dongming and Yu, Feng and Tian, Yueying and
               Tang, Jiaqi and Chen, Qing-Guo and Yang, Yang and Lu, Jianfeng},
  booktitle = {ICLR},
  year      = {2026},
  note      = {arXiv:2602.11743}
}

@inproceedings{zhang2022memo,
  title     = {{MEMO}: Test Time Robustness via Adaptation and Augmentation},
  author    = {Zhang, Marvin and Levine, Sergey and Finn, Chelsea},
  booktitle = {NeurIPS},
  year      = {2022}
}

@inproceedings{wang2021tent,
  title     = {Tent: Fully Test-Time Adaptation by Entropy Minimization},
  author    = {Wang, Dequan and Shelhamer, Evan and Liu, Shaoteng and
               Olshausen, Bruno and Darrell, Trevor},
  booktitle = {ICLR},
  year      = {2021},
  note      = {arXiv:2006.10726}
}

@inproceedings{liang2020shot,
  title     = {Do We Really Need to Access the Source Data?
               Source Hypothesis Transfer for Unsupervised Domain Adaptation},
  author    = {Liang, Jian and Hu, Dapeng and Feng, Jiashi},
  booktitle = {ICML},
  year      = {2020},
  note      = {arXiv:2002.08546}
}

@inproceedings{grandvalet2004entropy,
  title     = {Semi-Supervised Learning by Entropy Minimization},
  author    = {Grandvalet, Yves and Bengio, Yoshua},
  booktitle = {NeurIPS},
  year      = {2004}
}

@inproceedings{feng2023difftpt,
  title     = {Diverse Data Augmentation with Diffusions for Effective Test-time
               Prompt Tuning},
  author    = {Feng, Chun-Mei and Yu, Kai and Liu, Yong and Khan, Salman and Zuo, Wangmeng},
  booktitle = {ICCV},
  year      = {2023}
}

@inproceedings{yoon2024ctpt,
  title     = {C-TPT: Calibrated Test-Time Prompt Tuning for Vision-Language Models
               via Text Feature Dispersion},
  author    = {Yoon, Hee Suk and Yoon, Eunseop and Tee, Joshua Tian Jin and
               Hasegawa-Johnson, Mark and Li, Yingzhen and Yoo, Chang D.},
  booktitle = {ICLR},
  year      = {2024},
  note      = {arXiv:2403.14119}
}

@inproceedings{karmanov2024tda,
  title     = {Efficient Test-Time Adaptation of Vision-Language Models},
  author    = {Karmanov, Adilbek and Guan, Dayan and Lu, Shijian and
               El Saddik, Abdulmotaleb and Xing, Eric},
  booktitle = {CVPR},
  year      = {2024}
}

@inproceedings{zhu2024frolic,
  title     = {Enhancing Zero-Shot Vision Models by Label-Free Prompt Distribution
               Learning and Bias Correcting},
  author    = {Zhu, Xingyu and Zhu, Beier and Tan, Yi and Wang, Shuo and
               Hao, Yanbin and Zhang, Hanwang},
  booktitle = {NeurIPS},
  year      = {2024}
}

@article{zhou2022coop,
  title   = {Learning to Prompt for Vision-Language Models},
  author  = {Zhou, Kaiyang and Yang, Jingkang and Loy, Chen Change and Liu, Ziwei},
  journal = {International Journal of Computer Vision},
  volume  = {130},
  number  = {9},
  pages   = {2337--2348},
  year    = {2022}
}

@article{tsallis1988,
  title   = {Possible Generalization of Boltzmann-Gibbs Statistics},
  author  = {Tsallis, Constantino},
  journal = {Journal of Statistical Physics},
  volume  = {52},
  number  = {1--2},
  pages   = {479--487},
  year    = {1988}
}

@article{hill1973,
  title   = {Diversity and Evenness: A Unifying Notation and Its Consequences},
  author  = {Hill, M. O.},
  journal = {Ecology},
  volume  = {54},
  number  = {2},
  pages   = {427--432},
  year    = {1973}
}

@inproceedings{krogh1994ambiguity,
  title     = {Neural Network Ensembles, Cross Validation, and Active Learning},
  author    = {Krogh, Anders and Vedelsby, Jesper},
  booktitle = {NeurIPS},
  year      = {1994}
}

@inproceedings{balles2018adam,
  title     = {Dissecting Adam: The Sign, Magnitude and Variance of Stochastic Gradients},
  author    = {Balles, Lukas and Hennig, Philipp},
  booktitle = {ICML},
  year      = {2018},
  note      = {arXiv:1705.07774}
}

@inproceedings{loshchilov2019adamw,
  title     = {Decoupled Weight Decay Regularization},
  author    = {Loshchilov, Ilya and Hutter, Frank},
  booktitle = {ICLR},
  year      = {2019}
}

@article{mcnemar1947,
  title   = {Note on the Sampling Error of the Difference Between Correlated
             Proportions or Percentages},
  author  = {McNemar, Quinn},
  journal = {Psychometrika},
  volume  = {12},
  number  = {2},
  pages   = {153--157},
  year    = {1947}
}

@inproceedings{hendrycks2020augmix,
  title     = {AugMix: A Simple Data Processing Method to Improve Robustness
               and Uncertainty},
  author    = {Hendrycks, Dan and Mu, Norman and Cubuk, Ekin Dogus and Zoph, Barret and
               Gilmer, Justin and Lakshminarayanan, Balaji},
  booktitle = {ICLR},
  year      = {2020}
}

@inproceedings{deng2009imagenet,
  title     = {ImageNet: A Large-Scale Hierarchical Image Database},
  author    = {Deng, Jia and Dong, Wei and Socher, Richard and Li, Li-Jia and
               Li, Kai and Fei-Fei, Li},
  booktitle = {CVPR},
  year      = {2009}
}

@inproceedings{hendrycks2021nae,
  title     = {Natural Adversarial Examples},
  author    = {Hendrycks, Dan and Zhao, Kevin and Basart, Steven and
               Steinhardt, Jacob and Song, Dawn},
  booktitle = {CVPR},
  pages     = {15262--15271},
  year      = {2021}
}

@inproceedings{hendrycks2021manyfaces,
  title     = {The Many Faces of Robustness: A Critical Analysis of
               Out-of-Distribution Generalization},
  author    = {Hendrycks, Dan and Basart, Steven and Mu, Norman and Kadavath, Saurav and
               Wang, Frank and Dorundo, Evan and Desai, Rahul and Zhu, Tyler and
               Parajuli, Samyak and Guo, Mike and Song, Dawn and Steinhardt, Jacob and
               Gilmer, Justin},
  booktitle = {ICCV},
  year      = {2021}
}

@inproceedings{wang2019sketch,
  title     = {Learning Robust Global Representations by Penalizing Local
               Predictive Power},
  author    = {Wang, Haohan and Ge, Songwei and Lipton, Zachary and Xing, Eric P.},
  booktitle = {NeurIPS},
  year      = {2019}
}

@inproceedings{recht2019imagenetv2,
  title     = {Do ImageNet Classifiers Generalize to ImageNet?},
  author    = {Recht, Benjamin and Roelofs, Rebecca and Schmidt, Ludwig and
               Shankar, Vaishaal},
  booktitle = {ICML},
  year      = {2019}
}

@inproceedings{nilsback2008flowers,
  title     = {Automated Flower Classification over a Large Number of Classes},
  author    = {Nilsback, Maria-Elena and Zisserman, Andrew},
  booktitle = {Indian Conference on Computer Vision, Graphics and Image Processing},
  year      = {2008}
}

@inproceedings{cimpoi2014dtd,
  title     = {Describing Textures in the Wild},
  author    = {Cimpoi, M. and Maji, S. and Kokkinos, I. and Mohamed, S. and Vedaldi, A.},
  booktitle = {CVPR},
  year      = {2014}
}

@inproceedings{parkhi2012pets,
  title     = {Cats and Dogs},
  author    = {Parkhi, Omkar M. and Vedaldi, Andrea and Zisserman, Andrew and
               Jawahar, C. V.},
  booktitle = {CVPR},
  year      = {2012}
}

@inproceedings{krause2013cars,
  title     = {3D Object Representations for Fine-Grained Categorization},
  author    = {Krause, Jonathan and Stark, Michael and Deng, Jia and Fei-Fei, Li},
  booktitle = {ICCV Workshops},
  year      = {2013}
}

@article{soomro2012ucf101,
  title   = {UCF101: A Dataset of 101 Human Actions Classes from Videos in the Wild},
  author  = {Soomro, Khurram and Zamir, Amir Roshan and Shah, Mubarak},
  journal = {CoRR},
  volume  = {abs/1212.0402},
  year    = {2012}
}

@inproceedings{feifei2004caltech101,
  title     = {Learning Generative Visual Models from Few Training Examples:
               An Incremental Bayesian Approach Tested on 101 Object Categories},
  author    = {Fei-Fei, Li and Fergus, Rob and Perona, Pietro},
  booktitle = {CVPR Workshops},
  year      = {2004}
}

@inproceedings{bossard2014food101,
  title     = {Food-101 -- Mining Discriminative Components with Random Forests},
  author    = {Bossard, Lukas and Guillaumin, Matthieu and Van Gool, Luc},
  booktitle = {ECCV},
  year      = {2014}
}

@inproceedings{xiao2010sun397,
  title     = {SUN Database: Large-Scale Scene Recognition from Abbey to Zoo},
  author    = {Xiao, J. and Hays, J. and Ehinger, K. A. and Oliva, A. and Torralba, A.},
  booktitle = {CVPR},
  year      = {2010}
}

@techreport{maji2013aircraft,
  title       = {Fine-Grained Visual Classification of Aircraft},
  author      = {Maji, S. and Kannala, J. and Rahtu, E. and Blaschko, M. and Vedaldi, A.},
  institution = {arXiv},
  year        = {2013}
}

@article{helber2019eurosat,
  title   = {EuroSAT: A Novel Dataset and Deep Learning Benchmark for Land Use
             and Land Cover Classification},
  author  = {Helber, Patrick and Bischke, Benjamin and Dengel, Andreas and Borth, Damian},
  journal = {IEEE Journal of Selected Topics in Applied Earth Observations and
             Remote Sensing},
  year    = {2019}
}

@article{huang2026whatdrives,
  title   = {What Drives Test-Time Adaptation for CLIP? A Controlled Empirical Study
             from an Update Perspective},
  author  = {Huang, Jiazhen and Chen, Xiao and Liu, Zhiming and Sun, Yaru and
             Jiang, Jingyan and Wang, Zhi},
  journal = {arXiv preprint arXiv:2606.14299},
  year    = {2026}
}

@article{chaimanowong2024sorting,
  title   = {Approximating Invariant Functions with the Sorting Trick is
             Theoretically Justified},
  author  = {Chaimanowong, Wee and Zhu, Ying},
  journal = {arXiv preprint arXiv:2403.01671},
  year    = {2024}
}

@inproceedings{roy2007effrank,
  title     = {The Effective Rank: A Measure of Effective Dimensionality},
  author    = {Roy, Olivier and Vetterli, Martin},
  booktitle = {European Signal Processing Conference (EUSIPCO)},
  pages     = {606--610},
  year      = {2007}
}

@book{asmussen2010ruin,
  title     = {Ruin Probabilities},
  author    = {Asmussen, S{\o}ren and Albrecher, Hansj{\"o}rg},
  edition   = {2nd},
  publisher = {World Scientific},
  series    = {Advanced Series on Statistical Science \& Applied Probability},
  year      = {2010}
}

@article{chow1970,
  title   = {On Optimum Recognition Error and Reject Tradeoff},
  author  = {Chow, C. K.},
  journal = {IEEE Transactions on Information Theory},
  volume  = {16},
  number  = {1},
  pages   = {41--46},
  year    = {1970}
}

@inproceedings{jacot2018ntk,
  title     = {Neural Tangent Kernel: Convergence and Generalization in Neural Networks},
  author    = {Jacot, Arthur and Gabriel, Franck and Hongler, Cl{\'e}ment},
  booktitle = {NeurIPS},
  year      = {2018}
}

@inproceedings{zanella2024mta,
  title     = {On the Test-Time Zero-Shot Generalization of Vision-Language Models:
               Do We Really Need Prompt Learning?},
  author    = {Zanella, Maxime and Ben Ayed, Ismail},
  booktitle = {CVPR},
  year      = {2024}
}

@inproceedings{sheng2025rtpt,
  title     = {{R-TPT}: Improving Adversarial Robustness of Vision-Language Models
               through Test-Time Prompt Tuning},
  author    = {Sheng, Lijun and Liang, Jian and Wang, Zilei and He, Ran},
  booktitle = {CVPR},
  year      = {2025}
}

@inproceedings{zhao2024rlcf,
  title     = {Test-Time Adaptation with {CLIP} Reward for Zero-Shot Generalization
               in Vision-Language Models},
  author    = {Zhao, Shuai and Wang, Xiaohan and Zhu, Linchao and Yang, Yi},
  booktitle = {ICLR},
  year      = {2024}
}

@inproceedings{shanmugam2021tta,
  title     = {Better Aggregation in Test-Time Augmentation},
  author    = {Shanmugam, Divya and Blalock, Davis and Balakrishnan, Guha and Guttag, John},
  booktitle = {ICCV},
  year      = {2021}
}

@inproceedings{ahamed2026tpt,
  title     = {{A-TPT}: Angular Diversity Calibration Properties for Test-Time Prompt
               Tuning of Vision-Language Models},
  author    = {Ahamed, Shihab and Miriya Thanthrige, Udaya Sampath K. Perera and
               Rodrigo, Ranga and Khan, Muhammad Haris},
  booktitle = {ICLR},
  pages     = {69175--69199},
  year      = {2026}
}

@inproceedings{anonymous2026agitpt,
  title     = {{AGI-TPT}: Adaptive Geometry-Informed Test-Time Prompt Tuning for
               Calibrated Zero-Shot Vision-Language Models},
  author    = {Anonymous},
  booktitle = {Submitted to the Fifteenth International Conference on Learning
               Representations},
  year      = {2026},
  url       = {https://openreview.net/forum?id=dgirdRlD7Y},
  note      = {under review}
}

@article{ahamed2026m,
  title   = {{M-PFT}: Angular Margin for Calibrated Prompt-Tuning of Fine-Tuned
             Vision-Language Models},
  author  = {Ahamed, Shihab Aaqil and Ansaff, Mohamed and Phyu, Nang Khin Myat Lay and
             Thanthrige, Udaya Sampath K. and Edussooriya, Chamira U. S. and
             Rodrigo, Ranga and Khan, Muhammad Haris},
  journal = {arXiv preprint arXiv:2610.39861},
  year    = {2026}
}

@inproceedings{sharifdeen2026mvc,
  title     = {{MVC-Bench}: Benchmarking Calibration of Medical Vision-Language Models},
  author    = {Sharifdeen, Ashshak and Ahamed, Shihab Aaqil and Khan, Ufaq and
               Munir, Muhammad Akhtar and Ibrahim, Sujair and Ahamed, Mohamed Rafeek Mareer and
               Xie, Yutong and Razzak, Imran and Khan, Muhammad Haris},
  booktitle = {Findings of the Association for Computational Linguistics: EMNLP 2026},
  pages     = {36128--36162},
  year      = {2026}
}

@article{lin1991jsd,
  title   = {Divergence Measures Based on the Shannon Entropy},
  author  = {Lin, Jianhua},
  journal = {IEEE Transactions on Information Theory},
  volume  = {37},
  number  = {1},
  pages   = {145--151},
  year    = {1991}
}

\appendix
\section{Proofs}
\label{app:proofs}

All identities were verified against autograd in float64; the residual quoted after each
is the maximum absolute disagreement observed. Provenance is stated explicitly: results
that are classical are cited and not claimed.

\subsection{Gradient of the averaged-prediction entropy (\cref{eq:thmA})}
\label{app:gradient}
With $p_i=\mathrm{softmax}(z_i)$ and $L=H(\pbar)$, $\pbar=\frac1N\sum_i p_i$, the softmax
Jacobian $\partial p_{jc}/\partial z_{jc'} = p_{jc}(\delta_{cc'}-p_{jc'})$ gives
\[
\frac{\partial L}{\partial z_{jc}}
= -\frac1N p_{jc}\Big(\log\pbar_c - \sum_{c'}p_{jc'}\log\pbar_{c'}\Big)
= -\frac{1}{N}\sum_{c'} p_{jc}p_{jc'}\log\frac{\pbar_c}{\pbar_{c'}},
\]
using $\sum_{c'}p_{jc'}=1$ to write the bracket as a $p_j$-average of log-ratios.
Residual $2.8\times10^{-17}$ (first form) and $6.9\times10^{-17}$ (pairwise form).

\paragraph{Consequences.} \emph{(i)} The pair $(c,c')$ is gated by $p_{jc}p_{jc'}$, which
vanishes unless view $j$ places mass on both. \emph{(ii)} The sign of the force
$-\partial L/\partial z_{jc}$ is $\mathrm{sign}(\pbar_c-\pbar_{c'})$, a rank statistic of
$\pbar$ alone, verified with $0$ violations over $5220$ pairs. \emph{(iii)} a one-hot
view has $p_{jc}p_{jc'}=0$ for all $c\ne c'$ and contributes exactly zero.

\subsection{Why the analysis holds for any tuned parameters}
\label{app:reach}
For a loss $L$ depending on $\theta$ only through the logits $z$, the chain rule gives
$\nabla_\theta L = J^\top \nabla_z L$ with $J=\partial z/\partial\theta$. This requires only
that $z(\cdot)$ be differentiable at $\theta$: no assumption on width, initialisation,
architecture or step size. Hence $\nabla_z L$ is a factor of every parameter-space gradient,
and in particular $\nabla_z L = 0 \Rightarrow \nabla_\theta L = 0$ for any network. Verified on
a nonlinear encoder: $\lVert\nabla_\theta L - J^\top\nabla_z L\rVert_\infty = 2.2\times10^{-16}$.
This is the standard functional-gradient factorisation underlying kernel analyses of training
dynamics \citep{jacot2018ntk}; we use only the factorisation, not any infinite-width limit.

\subsection{The statistic $\Psi$ through which views reach the parameters}
\label{app:psi}
With $z_{ic}=s\langle v_i,w_c(\theta)\rangle$ and the view embeddings $v_i$ constant, write
$\Phi_{ic}=-\partial L/\partial z_{ic}$ for the \emph{force} on logit $z_{ic}$. This is the
sign convention used in \cref{eq:thmB} and \cref{app:cramer}, and it is the one under which
every ``positive iff'' statement in this appendix was verified; the singular values of $\Psi$
below are unaffected by the choice. Then
\[
\nabla_\theta L \;=\; s\sum_c \Big(\frac{\partial w_c}{\partial\theta}\Big)^{\!\top}\Psi_c,
\qquad \Psi := \Phi^\top V \in \mathbb{R}^{K\times D},
\]
so the entire dependence of the update on the views passes through $\Psi$. Being a sum of
$|S|$ outer products, $\operatorname{rank}\Psi\le|S|$ regardless of $K$ or $D$. Verified to
$3.3\times10^{-16}$. Two different view sets with the same $\Psi$ produce identical updates for
every encoder. We use ``sufficient statistic'' here in the plain sense that the map factors
through $\Psi$; this is not Fisher--Neyman statistical sufficiency.

\subsection{The boosting threshold}
From \cref{eq:thmA}, the force on $z_{jc}$ is positive iff $\pbar_c > \exp\langle p_j,\log\pbar\rangle$.
(This is the sign of $\nabla_z L$. It is not a claim about realised logit motion, which is
$-\eta\,JJ^\top\nabla_z L$ and need not share it coordinatewise.) For a single view
this reduces to $p_c > e^{-H(p)}$, the inverse perplexity: entropy minimisation boosts
every class above the inverse of the effective class count, not merely the argmax. The
$q$-deformed form replaces $e^{H}$ by the Hill number $D_q=(\sum_c p_c^q)^{1/(1-q)}$
\citep{hill1973}. Elementary given the Jacobian; $0$ violations.

\subsection{The dissent threshold is a Cram\'er root (\cref{eq:thmB})}
\label{app:cramer}
Let $b$ be a class view $j$ is confident in with $\pbar_b<\pbar_a$, $a$ the consensus
argmax. Put $L_c=\log(\pbar_c/\pbar_b)$, let $\tilde p_j$ be $p_j$ conditioned off $b$,
and $M_j(s)=\E_{c\sim\tilde p_j}[e^{sL_c}]$. Writing the objective in Tsallis form of
order $q$ \citep{tsallis1988} and splitting off the $c=b$ term,
\[
\Phi_{jb}>0 \iff
\begin{cases} M_j(q-1)<1, & q>1,\\ \E_{\tilde p_j}[L]<0, & q=1,\\ M_j(q-1)>1, & q<1.\end{cases}
\]
$M_j$ is strictly convex with $M_j(0)=1$, $M_j'(0)=\E_{\tilde p_j}[L]$, and
$M_j(s)\to\infty$ since $L_a>0$. Hence when $\E_{\tilde p_j}[L]<0$ there is a unique
$s^\star>0$ with $M_j(s^\star)=1$, and $q^\star = 1+s^\star$ is the exact per-view
threshold. Verified on $279$ dissenting views at $q\in\{1,1.2,1.5,2,3,5\}$, so $1674$
checks in total: the criterion reproduces the true sign with $0$ disagreements; $M_j(0)=1$ to $4.4\times10^{-16}$.

The equation $\log M(s)=0$ is the defining equation of the adjustment (Lundberg)
coefficient in large-deviations and ruin theory \citep{asmussen2010ruin}; the classical object
is not ours, its identification as the anti-consensus threshold of this objective is. We use it
as a characterisation, not as a tuning rule: $q^\star$ is a per-view quantity and we neither
estimate it online nor build a method on it.

\subsection{Agreement-blindness, and a step that increases disagreement}
$H(\pbar)$ depends on the views only through $\pbar$, so it is exactly constant on the
$(N-1)(K-1)$-dimensional slice $\{(p_i):\text{mean}=\pbar\}$, which contains both perfect
consensus and, for uniform $\pbar$ with $N=K$, maximal disagreement. Using the ambiguity
decomposition $H(\pbar)=\E_i[H(p_i)]+I(C;V)$, classical and due to \citet{lin1991jsd} in this
form, with the squared-error antecedent of \citet{krogh1994ambiguity}, and
equal to the Jensen--Shannon divergence of the views, we exhibit ($N=3$, $K=4$,
$\eta=0.05$) a descent step with $H(\pbar): 1.071359\to1.069901$ while
$I(C;V): 0.370935\to0.371380$. The loss reports progress while the views disagree
\emph{more}, contradicting the stated motivation of the objective. The honest caveat is
that $H(\pbar)\to0$ does force both terms to zero; the failure is that one scale-free
step at large $K$ never approaches that regime.

\section{The view-score audit and the selection ceiling}
\label{app:ceiling}

\begin{table}[h]
  \centering\small
  \caption{\textbf{The entropy score is at chance exactly where a better choice would help.}
  Mean AUC over six datasets per encoder, where $0.5$ means the score orders correct and
  incorrect views no better than chance. Orientation matters: a value below $0.5$ is
  anti-predictive, and reversing the image-feature row gives $0.546$ and $0.532$. That row tests
  statistics computed from the cached embeddings rather than the output distribution, and the
  crop-area row a quantity that is not a function of the model at all.}
  \label{tab:blind}
  \vspace{2pt}
  \begin{tabular}{lcc}
    \toprule
    Score, and images it is evaluated on & ViT-B/16 & RN50 \\
    \midrule
    Entropy, all images & $0.788$ & $0.738$ \\
    \textbf{Entropy, reducible errors (controlled)} & $\mathbf{0.503}$ & $\mathbf{0.501}$ \\
    Entropy, reducible errors (uncontrolled) & $0.504$ & $0.503$ \\
    \midrule
    Image-feature statistics, reducible errors & $0.454$ & $0.468$ \\
    Crop area, reducible errors & $\mathit{0.510}$ & --- \\
    \bottomrule
  \end{tabular}
\end{table}

\begin{table}[h]
  \centering\small
  \begin{tabular}{@{}llrrr@{}}
    \toprule
    Dataset & Backbone & Achieved & Oracle & Headroom \\
    \midrule
    ImageNet-A & ViT-B/16 & $61.5$ & $83.4$ & $+21.8$ \\
    ImageNet-A & RN50 & $32.4$ & $59.2$ & $\mathbf{+26.8}$ \\
    ImageNet & ViT-B/16 & $69.6$ & $85.6$ & $+16.0$ \\
    ImageNet & RN50 & $59.8$ & $78.0$ & $+18.2$ \\
    ImageNet-R & ViT-B/16 & $77.8$ & $90.2$ & $+12.4$ \\
    DTD & ViT-B/16 & $46.1$ & $64.7$ & $+18.6$ \\
    Flowers102 & ViT-B/16 & $65.3$ & $79.1$ & $+13.9$ \\
    \bottomrule
  \end{tabular}
  \caption{\textbf{How much a better choice of views could buy.} The oracle keeps the
  same number of views as the deployed method but selects them knowing the label.}
  \label{tab:ceiling}
\end{table}

\begin{table}[h]
  \centering\small
  \caption{\textbf{Closed forms agree with automatic differentiation at machine precision.}
  Maximum absolute disagreement, float64 (machine epsilon $\approx 2.2\times10^{-16}$). An
  algebra slip would produce an order-one disagreement, not one at this scale.}
  \label{tab:resid}
  \vspace{2pt}
  \begin{tabular}{@{}lc@{}}
    \toprule
    Identity & Residual \\
    \midrule
    Logit-space gradient of $H$ & $6.9\times10^{-17}$ \\
    \Cref{eq:thmA}, first form & $3.5\times10^{-17}$ \\
    \Cref{eq:thmA}, pairwise form & $2.1\times10^{-17}$ \\
    Hessian of $H\circ\mathrm{softmax}$ & $8.3\times10^{-17}$ \\
    Tangent eigenvalues at the uniform point ($=-1/K$) & $1\times10^{-16}$ \\
    \midrule
    Suppression peak at $e^{-(H+1)}$ & 5 s.f. \\
    Argmax invariance under gradient flow (reproduces \citealp{farina2024zero}, Prop.~2.1) & 0 flips / 3000 steps \\
    \Cref{eq:thmB} sign criterion, all three cases & 0 / 1674 \\
    \bottomrule
  \end{tabular}
\end{table}

\section{The optimiser and the channel}
\label{app:optimiser}

That Adam's update direction is the \emph{sign} of the gradient is established
\citep{balles2018adam}; we claim no novelty for the identity, only for its consequence under a
per-sample optimiser reset, which makes every update by AdamW \citep{loshchilov2019adamw}
a first step from zero moments. From
$\hat m = g$, $\hat v = g^2$ the update is $-\eta\,g/(|g|+\varepsilon)\to-\eta\,\mathrm{sign}(g)$,
so $\|\Delta\theta\|_2=\eta\sqrt{P}$ independently of $\|g\|$. Verified to $1.9\times10^{-4}$ on
a $0.226$-norm update; exact only while $|g_i|\gg\varepsilon$, and under fp16 an underflowed
coordinate takes a step of exactly zero. The direction fares no better:
$\cos(-g,\Delta\theta)=\|g\|_1/(\sqrt P\|g\|_2)$ is $0.798$ for a dense Gaussian gradient but
$0.028$ measured on CLIP's heavy-tailed prompt gradient.

Because the view features are constants, the image enters the update only through
$\Psi := \Phi^\top V \in \mathbb{R}^{K \times D}$, a sum of $|S|$ outer products, so
$\operatorname{rank}\Psi \le |S|$ regardless of $K$ or $D$. Its singular values have
participation ratio $(\sum_i\sigma_i)^2/\sum_i\sigma_i^2$ with median $1.08$ (ImageNet-A,
ViT-B/16, $n=600$). Participation ratio is one of two standard conventions; the entropy
effective rank \citep{roy2007effrank} gives $1.23$ on the same $\Psi$. The two differ
materially, so we name which we report.

\section{The matched-information control in full}
\label{app:matched}

\begin{figure}[h]
  \centering
  \includegraphics[width=\linewidth]{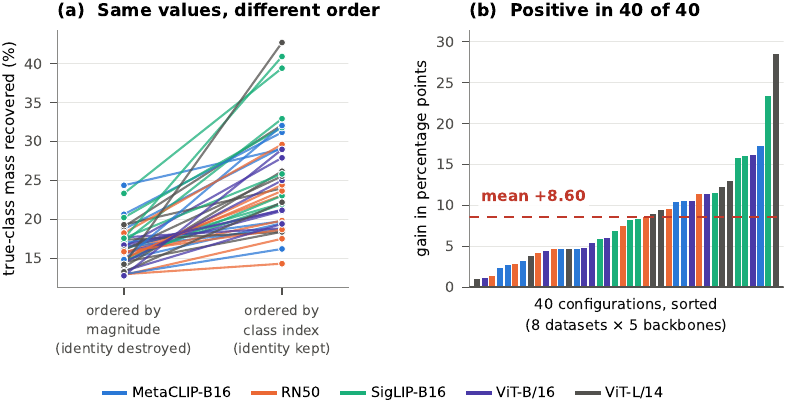}
  \caption{\textbf{What the model's response cannot see.} Both arms use the same regressor, the
  same training rows, and feature sets of identical dimension holding the identical value
  multiset; only the ordering differs. \textbf{(a)} Every configuration improves when class
  identity is preserved. \textbf{(b)} The gap, sorted: positive in $40/40$, mean $8.60$ percentage points.
  The capacity ladder had saturated in all $40$, so under-fitting does not explain it.
  Per-configuration numbers in \cref{tab:matched}.}
  \label{fig:matched}
\end{figure}

The standard confidence functionals, Shannon, Tsallis and R\'enyi at fixed order,
max-probability, top-$2$ margin, are invariant to permuting the class labels, hence
functions of the \emph{sorted} probability vector alone (verified to $3.6\times10^{-15}$;
the factorisation itself is classical, \citealp{chaimanowong2024sorting}). Such a score can measure
how concentrated a view is, never \emph{which} class it is concentrated on. To measure
what that costs, we fit the same regressor, on the same rows, to predict a view's
true-class mass from two feature sets of \emph{identical dimension and identical value
multiset}, differing only in ordering: the full sorted spectrum (class identity
destroyed) versus the log-probability vector (class identity preserved). Any gap cannot
be dimensionality or capacity.

Across $8$ datasets $\times$ $5$ backbones ($40$ configurations), preserving class
identity is worth $8.60$ percentage points, positive in $40/40$, with the capacity ladder saturated in
all $40$ (\cref{fig:matched}, \cref{tab:matched}). In share terms, the best class-symmetric fit captures
$8.4\%$ of the oracle gap against $21.7\%$ for the class-asymmetric one. This is an
information measurement, not a method, both arms are fit with labels, and we note
that the one deployed method which does break this symmetry, ADTE's class-specific
exponent \citep{wu2026adte}, buys only $\approx 0.5$ percentage points end to end in its own ablation.
The information is present; extracting it label-free is unsolved.

Both arms use the identical regressor, the identical training rows, the identical
readout, and feature sets of identical dimension $K$ holding the identical value
multiset, they differ only in whether the entries are ordered by magnitude (class
identity destroyed) or by class index (class identity preserved). A capacity ladder
(\text{max\_iter} $\in\{500,2000\}$) was run on the symmetric arm and had saturated in
$40/40$ configurations, so the gap is not an under-fitting artefact. Both arms are fit
against the true-class mass \emph{with labels}: this measures information content, not a
deployable method.

\begin{table}[h]
  \centering\footnotesize
  \begin{tabular}{@{}llrrrrrrr@{}}
    \toprule
    Backbone & Dataset & $K$ & $n$ & Entropy & Sorted & Unsorted & Oracle & $\Delta$ \\
    \midrule
    MetaCLIP-B16 & ImageNet-A & 200 & 496 & $19.96$ & $18.75$ & $29.23$ & $79.64$ & $\mathbf{+10.48}$ \\
    MetaCLIP-B16 & ImageNet & 1000 & 359 & $19.22$ & $16.99$ & $19.78$ & $77.99$ & $\mathbf{+2.79}$ \\
    MetaCLIP-B16 & IN-Sketch & 1000 & 389 & $16.71$ & $16.45$ & $18.77$ & $69.67$ & $\mathbf{+2.31}$ \\
    MetaCLIP-B16 & IN-V2 & 1000 & 405 & $13.09$ & $14.81$ & $19.51$ & $78.77$ & $\mathbf{+4.69}$ \\
    MetaCLIP-B16 & StanfordCars & 196 & 279 & $24.73$ & $24.37$ & $29.03$ & $79.93$ & $\mathbf{+4.66}$ \\
    MetaCLIP-B16 & Food101 & 101 & 247 & $21.46$ & $20.65$ & $31.17$ & $84.21$ & $\mathbf{+10.53}$ \\
    MetaCLIP-B16 & SUN397 & 397 & 432 & $13.89$ & $12.96$ & $16.20$ & $78.94$ & $\mathbf{+3.24}$ \\
    MetaCLIP-B16 & UCF101 & 101 & 337 & $15.43$ & $14.84$ & $32.05$ & $76.26$ & $\mathbf{+17.21}$ \\
    RN50 & ImageNet-A & 200 & 658 & $15.81$ & $15.65$ & $23.10$ & $78.88$ & $\mathbf{+7.45}$ \\
    RN50 & ImageNet & 1000 & 451 & $12.86$ & $12.86$ & $17.52$ & $79.16$ & $\mathbf{+4.66}$ \\
    RN50 & IN-Sketch & 1000 & 519 & $16.38$ & $15.61$ & $19.85$ & $73.22$ & $\mathbf{+4.24}$ \\
    RN50 & IN-V2 & 1000 & 517 & $12.96$ & $12.96$ & $14.31$ & $76.60$ & $\mathbf{+1.35}$ \\
    RN50 & StanfordCars & 196 & 613 & $15.17$ & $15.01$ & $23.65$ & $79.28$ & $\mathbf{+8.65}$ \\
    RN50 & Food101 & 101 & 356 & $14.61$ & $14.89$ & $24.44$ & $82.87$ & $\mathbf{+9.55}$ \\
    RN50 & SUN397 & 397 & 492 & $15.65$ & $15.85$ & $18.70$ & $79.47$ & $\mathbf{+2.85}$ \\
    RN50 & UCF101 & 101 & 422 & $18.25$ & $18.25$ & $29.62$ & $78.44$ & $\mathbf{+11.37}$ \\
    SigLIP-B16 & ImageNet-A & 200 & 510 & $20.78$ & $23.33$ & $39.41$ & $86.27$ & $\mathbf{+16.08}$ \\
    SigLIP-B16 & ImageNet & 1000 & 290 & $16.90$ & $16.21$ & $22.07$ & $89.31$ & $\mathbf{+5.86}$ \\
    SigLIP-B16 & IN-Sketch & 1000 & 319 & $15.05$ & $15.05$ & $21.94$ & $83.07$ & $\mathbf{+6.90}$ \\
    SigLIP-B16 & IN-V2 & 1000 & 382 & $16.49$ & $14.66$ & $23.04$ & $89.01$ & $\mathbf{+8.38}$ \\
    SigLIP-B16 & StanfordCars & 196 & 164 & $18.90$ & $17.07$ & $32.93$ & $87.20$ & $\mathbf{+15.85}$ \\
    SigLIP-B16 & Food101 & 101 & 173 & $19.08$ & $20.23$ & $31.79$ & $89.60$ & $\mathbf{+11.56}$ \\
    SigLIP-B16 & SUN397 & 397 & 449 & $17.59$ & $17.59$ & $25.84$ & $85.30$ & $\mathbf{+8.24}$ \\
    SigLIP-B16 & UCF101 & 101 & 330 & $17.58$ & $17.58$ & $40.91$ & $86.06$ & $\mathbf{+23.33}$ \\
    ViT-B/16 & ImageNet-A & 200 & 534 & $17.23$ & $16.48$ & $27.90$ & $84.83$ & $\mathbf{+11.42}$ \\
    ViT-B/16 & ImageNet & 1000 & 372 & $15.05$ & $16.40$ & $21.24$ & $82.53$ & $\mathbf{+4.84}$ \\
    ViT-B/16 & IN-Sketch & 1000 & 484 & $15.29$ & $15.70$ & $21.07$ & $78.10$ & $\mathbf{+5.37}$ \\
    ViT-B/16 & IN-V2 & 1000 & 435 & $16.32$ & $17.70$ & $18.85$ & $80.46$ & $\mathbf{+1.15}$ \\
    ViT-B/16 & StanfordCars & 196 & 515 & $17.86$ & $16.70$ & $21.17$ & $80.19$ & $\mathbf{+4.47}$ \\
    ViT-B/16 & Food101 & 101 & 208 & $17.79$ & $14.42$ & $25.00$ & $86.06$ & $\mathbf{+10.58}$ \\
    ViT-B/16 & SUN397 & 397 & 479 & $13.36$ & $13.36$ & $19.42$ & $81.84$ & $\mathbf{+6.05}$ \\
    ViT-B/16 & UCF101 & 101 & 376 & $12.77$ & $12.77$ & $28.99$ & $76.33$ & $\mathbf{+16.22}$ \\
    ViT-L/14 & ImageNet-A & 200 & 299 & $20.07$ & $19.06$ & $32.11$ & $85.28$ & $\mathbf{+13.04}$ \\
    ViT-L/14 & ImageNet & 1000 & 321 & $19.31$ & $19.31$ & $23.99$ & $82.87$ & $\mathbf{+4.67}$ \\
    ViT-L/14 & IN-Sketch & 1000 & 413 & $16.71$ & $14.53$ & $18.40$ & $80.39$ & $\mathbf{+3.87}$ \\
    ViT-L/14 & IN-V2 & 1000 & 397 & $17.13$ & $17.38$ & $18.39$ & $82.12$ & $\mathbf{+1.01}$ \\
    ViT-L/14 & StanfordCars & 196 & 317 & $15.14$ & $13.88$ & $26.18$ & $82.97$ & $\mathbf{+12.30}$ \\
    ViT-L/14 & Food101 & 101 & 137 & $17.52$ & $16.06$ & $25.55$ & $89.05$ & $\mathbf{+9.49}$ \\
    ViT-L/14 & SUN397 & 397 & 392 & $11.99$ & $13.27$ & $22.19$ & $81.12$ & $\mathbf{+8.93}$ \\
    ViT-L/14 & UCF101 & 101 & 281 & $13.88$ & $14.23$ & $42.70$ & $84.34$ & $\mathbf{+28.47}$ \\
    \midrule
    \multicolumn{8}{@{}l}{\textit{mean over all 40 configurations}} & $\mathbf{+8.60}$ \\
    \bottomrule
  \end{tabular}
  \caption{\textbf{Keeping class identity is worth $8.60$ percentage points, positive in $40/40$.}
  ``Sorted'' and ``Unsorted'' are the two matched feature sets; ``Oracle'' ranks by the
  true-class mass directly. Share of the oracle gap: $8.4\%$ for the best
  class-symmetric fit, $21.7\%$ for the class-asymmetric one.}
  \label{tab:matched}
\end{table}

\section{Comparison to a state-of-the-art selection rule}
\label{app:adte}

ADTE \citep{wu2026adte} combines a class-adaptive Tsallis entropy for view selection
with a logit adjustment derived from a transductively estimated class prior
\citep{zhu2024frolic}. Only the selection component lies in our scope, and their own
ablation indicates that with logit adjustment held fixed the entropy variant accounts
for a minority of their margin. We implemented their selection rule from the paper's
equations at their published parameter interval. On our pipeline it underperforms
plain entropy on the suite average, which we attribute to a limitation of our setup
rather than of their method: we estimate the prior from $2000$ cached samples, which
on ImageNet is $2$ samples per class, whereas their procedure maintains a memory bank
over the full test stream. The degradation scales with the number of classes exactly
as that explanation predicts. We therefore report the per-sample Tsallis row of
\cref{tab:main} as the faithful comparison and cite their published system numbers
rather than a reimplementation we cannot validate.

\section{One measurement explains several separate results}
\label{app:explains}

The reducible AUC accounts for findings we had obtained separately and could not previously
connect. Selection is the largest positive stage because the AUC across all images is $0.79$; no
selector we constructed improved on entropy because the AUC on reducible errors is
approximately $0.50$; re-running selection on the model's own pseudo-labels recovers exactly
nothing because it is a fixed point by construction; and alternative aggregation rules give at
most $0.20$ percentage points because the view \emph{set}, not the pooling rule, is the binding
constraint. The finding also extends beyond prompt tuning: for MEMO \citep{zhang2022memo}, which
adapts model parameters rather than a prompt and performs no view selection, the gradient step
improves the averaged prediction in $0$ of $54$ configurations spanning two parameter groups and
nine learning rates over four orders of magnitude.

\section{Encoder generality in full}
\label{app:backbones}
On six datasets evaluated with both encoders, the mean gain is $1.18$ percentage points on RN50
with $4$ of $6$ datasets significant, against $0.82$ with $3$ of $6$ on ViT-B/16. Three datasets
are significant on both encoders independently. The relationship between the label-free
crop-area AUC and the realised gain holds within each encoder, with Spearman $\rho=0.943$ on
ViT-B/16 and $\rho=0.771$ on RN50, pooling to $\rho=0.841$ across all runs. The per-dataset
gains do not transfer across encoders: their rank correlation is $0.314$, which at this sample
size is indistinguishable from no relationship.

\section{Claims of ours that the controls refuted}
\label{app:failed}

\emph{(i)} ``The entropy score is \emph{uninformative} where the headroom lies.'' False. Under a
split-half design it reaches $20.15\%$ accuracy against $17.17\%$ for random selection, an
advantage of $2.98$ percentage points with a $95\%$ bootstrap interval of $[1.28, 4.72]$. The
defensible statement is quantitative: it captures about $4.5\%$ of what an oracle recovers.

\emph{(ii)} ``No label-free score beats entropy selection.'' False. That conclusion had been
drawn from a family of scores we constructed ourselves. A Tsallis score at fixed order $q=0.5$
beats Shannon entropy on $11$ of $15$ datasets, ties on one and loses on three, in the direction
of ADTE's own ablation \citep{wu2026adte}.

\emph{(iii)} ``The useful views are the dissenting ones.'' False. An oracle restricted to
dissenting views reaches $55.67\%$ against $59.73\%$ for a random half of the views.

\emph{(iv)} Two mechanisms for why crop area helps, each written down before the run. The first,
that area matters when the object fills the frame but not for textures, predicted that DTD would
not benefit; DTD gains $1.36$ percentage points on RN50, although it is the one dataset that
loses on ViT-B/16, and EuroSAT, predicted to be the least favourable case, shows the largest gain
in the suite. The second, that the driver is the upscale
factor, gives $\rho=0.348$ with $p=0.22$.

\emph{(v)} A stronger symmetry claim: that no class-symmetric score could exceed a ceiling we had
estimated. ADTE's class-specific exponent is not class-symmetric, so the family does not bound
the field, and our estimator never attained the ceiling it claimed. The matched-multiset
comparison of \cref{sec:audit} is the weaker and better-controlled replacement.

\section{Experimental design and the reasoning behind it}
\label{app:design}

This appendix states, for each analytical choice and each measurement in the paper, what question
it answers, which assumption it rests on, what alternative explanation it was built to exclude,
and why the chosen metric is the right one. We record the reasoning because the design, rather
than the size of any single number, is what makes the central claim defensible. The first four
subsections concern the analysis; the rest concern the experiments.

\subsection{Why the analysis is carried out in logit space}
\label{app:whylogit}
The methods in this family do not tune the same parameters. TPT updates four prompt token
embeddings; MEMO updates network parameters; a future variant could update something else again.
A statement proved about one parameter vector $\theta$ would therefore say nothing about the
family, and the family is what the paper is about. We avoid that by proving statements about the
derivative with respect to the \emph{logits} $z$, the pre-softmax class scores, and then
transporting them with the chain rule
\[
  \nabla_\theta L \;=\; J^\top \nabla_z L , \qquad J = \partial z/\partial\theta .
\]
This identity requires only that $z(\cdot)$ be differentiable at $\theta$. It assumes nothing
about width, initialisation, architecture, learning rate or step count, and it is exact rather
than asymptotic: we checked it on a nonlinear encoder to
$\lVert\nabla_\theta L - J^\top\nabla_z L\rVert_\infty = 2.2\times10^{-16}$
(\cref{app:reach}).

\paragraph{What we borrow from kernel analyses, and what we deliberately do not.}
Writing the update in function space rather than parameter space is the move that defines the
neural tangent kernel \citep{jacot2018ntk}. There the same factorisation is composed with itself
to give the function-space dynamics $\dot z = -\eta\,\Theta\,\nabla_z L$ with
$\Theta = J J^\top$, and the paper's theorem is that in the infinite-width limit, under the NTK
parameterisation with random initialisation and an infinitesimal learning rate, $\Theta$
converges to a deterministic kernel and stays \emph{constant} throughout training, so the
dynamics linearise. We use the factorisation. We do not use the theorem, and we are explicit
that we may not: our setting violates every one of its hypotheses. There are $P=2048$ prompt
parameters rather than infinite width; initialisation is a pretrained checkpoint at the embedding
of ``a photo of a'' rather than a random draw in the NTK parameterisation; and a single AdamW
step has $\lVert\Delta\theta\rVert_2 = \eta\sqrt{P} = 0.226$, roughly a quarter of a token
embedding's norm, which is neither infinitesimal nor lazy.

Rather than rest on that argument we measured the assumption directly. We form
$\Theta = J J^\top$ over $64$ logits, with $J$ taken through the real CLIP ViT-B/16 text encoder
and TPT's own prompt learner, take one step of the size the method actually takes, and recompute.
The relative change is $\lVert\Theta_1-\Theta_0\rVert_F/\lVert\Theta_0\rVert_F = 1.75$, with the
leading eigenvalue rising from $8.4\times10^{4}$ to $2.3\times10^{5}$. The kernel does not drift
slightly; it nearly triples over one step. Treating it as constant here would be false, so no NTK
convergence or generalisation conclusion is available to this paper and none is claimed.

\paragraph{What this licenses, and the line we do not cross.}
Two kinds of statement survive without any assumption on $\Theta$. First, exact zeros:
$\nabla_z L = 0 \Rightarrow \nabla_\theta L = 0$ for every Jacobian, which is what makes the
one-hot-view consequence of \cref{eq:thmA} parameterisation-free. Second, statements about the
\emph{content} of the signal, since $\nabla_z L$ is a factor of every parameter-space update
whatever $J$ is: if the only class-dependent information entering $\nabla_z L$ is the ordering of
$\pbar$, no choice of tuned parameters can recover a margin that was never in the factor.
Statements about realised logit \emph{motion} do not survive, because the realised change is
$-\eta\,\Theta\,\nabla_z L$ and $\Theta$ is far from a multiple of the identity: measured over
$96$ logits, $96\%$ of its Frobenius mass is off-diagonal and its condition number is $1838$, so
the per-coordinate sign agreement between $-\nabla_z L$ and the actual logit change is $61.5\%$
against a chance value of $50\%$. We therefore phrase \cref{eq:thmA,eq:thmB} as facts about the \emph{force}
$\partial L/\partial z_{jc}$ and never as claims about which logits rise, and where the paper
describes what the update does in practice (\cref{sec:theory}, final paragraph) the statement is
measured rather than derived.

\subsection{Why the view embeddings may be treated as constants}
The factorisation $\Psi = \Phi^\top V$ of \cref{app:psi}, through which the whole dependence of
the update on the views passes, needs the view embeddings $V$ to be constants. This is not an
approximation we impose: in the reference implementation the image encoder runs without gradient
tracking, so $V$ carries no gradient by construction and the identity is exact, verified to
$3.3\times10^{-16}$. The same fact is what makes the cached protocol below legitimate rather than
merely convenient, since a quantity that never receives a gradient can be computed once and
replayed.

\subsection{Why one step, and why the sign of the gradient}
Analysing a single update would be a simplification for ordinary training, but here it is a
description. The optimiser state is reset for every test image and one step is taken, so the
moment estimates are zero when the step is formed; Adam's first step from zero moments is the
sign of the gradient up to the numerical stabiliser \citep{balles2018adam}. The step we analyse
is therefore the step the method takes, agreeing with the implemented update to
$1.9\times10^{-4}$ at a step norm of $0.226$. This is also why the magnitude claim in
\cref{sec:theory} is a property of the deployed configuration rather than a worst case: an image
on which the model is nearly indifferent, where the gradient is small, receives exactly the same
sized perturbation as one on which it is certain.

\subsection{Why the results are stated for a family of scores rather than for Shannon entropy}
The published scores in this pipeline are Shannon entropy \citep{shu2022tpt}, a zero-temperature
average \citep{farina2024zero}, and Tsallis entropy with a class-adaptive exponent
\citep{wu2026adte}. Proving statements for the Tsallis family of order $q$ at once therefore
covers the variants that exist rather than only the first of them, and it costs nothing, because
the family recovers Shannon entropy as $q\to1$ \citep{tsallis1988}. It also reveals something a
Shannon-only analysis cannot: the condition under which the update pushes a view further from the
average has a single critical order $q^\star$, and Shannon entropy sits at the degenerate root
$q=1$ for \emph{every} distribution (\cref{app:cramer}). Separately, the permutation-symmetry
result is a statement about a function class, not an empirical average over methods, so it is
established by exhibiting the algebra and enumerating the class members rather than by
measurement, and ADTE is reported as the counterexample that keeps the class from being the whole
literature.

\subsection{Why a cached, fixed-view protocol}
Every arm of every comparison replays one cached tensor of view embeddings. Because the image
encoder runs without gradient tracking, the $64$ view embeddings per image are constants and can
be computed once. Two consequences follow. First, all arms see bit-identical views, so an
observed difference cannot come from augmentation randomness, and no paired test is contaminated
by resampling. Second, because only the selection rule varies across arms, the comparisons measure a property
of that pipeline stage rather than of any one method's optimiser. The entropy baseline is
evaluated at its published setting and is never tuned while our rule cross-fits two parameters;
we state that asymmetry rather than claim it favours the baseline.

\subsection{Why AUC for the view score, and not accuracy}
The question is whether a score \emph{orders} views correctly, not whether any particular
threshold on it is good. Accuracy of a downstream selector confounds the score with the choice of
how many views to keep, with the aggregation rule, and with the base difficulty of the dataset.
The area under the ROC curve is the probability that a randomly chosen correct view outscores a
randomly chosen incorrect one. It is threshold-free, invariant to any monotone rescaling of the
score, and has an unambiguous null value of $0.5$, so ``no better than chance'' is a statement
about the score itself rather than about a tuned pipeline. It is also computable per image with
no method in the loop, which is what allows it to be used as an \emph{a priori} predictor in
\cref{sec:method}.

\subsection{Why the population must be restricted, and the confound this creates}
A view score can only change an answer on images where a different choice of views yields a
different prediction. Averaged over all images, a score is dominated by cases where selection is
irrelevant because the model is already right, or hopeless because no view is right. Restricting
to reducible errors is therefore not a convenience: it is the only population on which the
quantity of interest is defined. The restriction introduces a dependence, because reducibility
is defined using the averaged prediction and confident views agree with one another. The
split-half design (\cref{fig:collapse}a) breaks that dependence by construction. We report both
the controlled and uncontrolled numbers so that the size of the confound is visible rather than
asserted to be small.

\subsection{Why the matched-multiset comparison, rather than an ablation}
\label{app:matched2}
To ask what class identity is worth, an ordinary ablation would compare a model given class
indices against one that is not, which changes the input dimension and the amount of information
simultaneously. Our two arms instead receive vectors of identical length containing the identical
multiset of numbers, differing only in whether they are sorted by magnitude or indexed by class.
Capacity, dimensionality and the values themselves are therefore held exactly fixed, and the only
free variable is whether the association between a value and its class survives. A capacity
ladder is run on the arm we expect to lose, because under-fitting that arm would inflate the
measured gap in our favour; the ladder saturates in all $40$ configurations.

\subsection{Why an oracle upper bound}
An oracle selector uses the ground-truth label and is not a method. Its role is to convert the
vague question ``could selection do better?'' into a number: it is the best achievable accuracy
for \emph{any} rule that selects the same number of views from the same candidates. Without it,
a small gain has no denominator, and the claim that a score captures $4.5\%$ of what is available
could not be stated.

\subsection{Why cross-fitting, and why the baseline is never tuned}
Our rule has two free parameters, $\omega$ and $m$; the baseline has none that we adjust. To keep
the comparison honest under that asymmetry we cross-fit: the data are divided into five parts and
the parameters used to score each part are selected using only the other four, so no image is
scored under parameters chosen with knowledge of its own outcome. The baseline is evaluated at
its published setting. The resulting comparison is therefore biased \emph{against} the proposed
rule, which is the direction an honest comparison should err in.

\subsection{Why McNemar, bootstrap intervals, and a disjoint-sample replication}
Predictions from two selection rules on the same image are paired, so an unpaired test would
overstate the variance and understate significance. McNemar's test conditions on exactly the
images where the two rules disagree, which is the relevant evidence. AUC has no simple closed-form
interval under this design, so we use bootstrap resampling over $2000$ replicates. For the
predictive relationship of \cref{sec:method}, correlating a measurement with a gain computed on
the same images would be circular; we therefore recompute the two quantities on disjoint halves
and report that correlation as well.

\subsection{Why MEMO, and what it does and does not test}
MEMO adapts model parameters rather than a prompt and performs no view selection. It is included
precisely because it differs in the parameter set: if the gradient step were inert only for a
particular parameterisation, MEMO would show a gain. Sweeping two parameter groups and nine
learning rates across four orders of magnitude, it does not. This tests whether the result is an
artefact of one parameter choice; it does not test architectures beyond those studied.

\subsection{Pre-registration, and predictions that failed}
Where a mechanism was proposed, the prediction was written down before the run. Three such
predictions were rejected: that the two selection signals would combine additively, and two
separate accounts of why crop area helps. We report them because a design that never contradicts
its author is not testing anything.

\section{Reproducibility}
\label{app:repro}

\paragraph{Benchmark composition.}
The $15$-dataset suite is ImageNet \citep{deng2009imagenet} with four distribution shifts,
ImageNet-A \citep{hendrycks2021nae}, ImageNet-R \citep{hendrycks2021manyfaces}, ImageNet-Sketch
\citep{wang2019sketch} and ImageNet-V2 \citep{recht2019imagenetv2}, together with Flowers102
\citep{nilsback2008flowers}, DTD \citep{cimpoi2014dtd}, OxfordPets \citep{parkhi2012pets},
StanfordCars \citep{krause2013cars}, UCF101 \citep{soomro2012ucf101}, Caltech101
\citep{feifei2004caltech101}, Food101 \citep{bossard2014food101}, SUN397 \citep{xiao2010sun397},
FGVC-Aircraft \citep{maji2013aircraft} and EuroSAT \citep{helber2019eurosat}.

All experiments replay a cached feature tensor: because the image tower runs under
\texttt{no\_grad}, the $64$ view features per sample are constants and can be computed
once. Every comparison in the paper therefore uses \emph{identical} views, cache and
readout, so only the quantity under study differs. Before any experiment we verified all
$15$ dataset loaders by reproducing published zero-shot accuracy at full test-set size
(all within tolerance, ImageNet to $0.01$ points). Significance is paired McNemar
\citep{mcnemar1947}; AUC confidence intervals are bootstrap over $2000$ resamples;
selector parameters are chosen by $5$-fold cross-fitting so that no sample is scored
under parameters chosen using it. Baselines are never tuned.

\section{Extended related work}
\label{app:related}

\paragraph{The incumbent line, in more detail.}
TPT \citep{shu2022tpt} fixed the recipe this paper studies: $64$ views per test image, a
confidence filter retaining the lowest-entropy $10\%$, an average over the survivors, and one
AdamW step on the prompt tokens minimising $H(\pbar)$. ZERO \citep{farina2024zero} keeps the
views and the filter, deletes the step, and replaces the average by a zero-temperature vote;
ADTE \citep{wu2026adte} keeps ZERO's pipeline and replaces the score by a class-adaptive Tsallis
entropy. Each states that it inherits the previous setup, which is why we treat the protocol as
shared rather than as TPT's. C-TPT \citep{yoon2024ctpt} and DiffTPT \citep{feng2023difftpt} vary
calibration and view generation respectively while leaving the scoring input untouched.

\paragraph{Calibration-oriented variants of TPT.}
C-TPT \citep{yoon2024ctpt}, A-TPT \citep{ahamed2026tpt} and AGI-TPT \citep{anonymous2026agitpt}
inherit TPT's views, confidence filter and prompt update, and add a regulariser on the
angular geometry of the class text features so that the averaged prediction is better
calibrated. AGI-TPT sets its separation target from the number of classes and the feature
dimension rather than tuning it, and is aimed at the online setting, where the prompt is
carried across test images and calibration otherwise drifts; M-PFT \citep{ahamed2026m} applies angular-margin regularisation to prompt-tuned
models more generally, and MVC-Bench \citep{sharifdeen2026mvc} benchmarks calibration of
medical vision-language models. Their evaluation metric is calibration error of the final
prediction, not the ranking of individual views, so their results and ours are not on a
common scale. Because they retain the entropy-based view filter, the audit of
\cref{sec:audit} applies to the views they select; whether a better-calibrated prompt also
raises the view-level AUC on reducible errors is an open question we do not test.

\paragraph{Why the second line does not escape the audit.}
Methods scoring views on the representation rather than the posterior
\citep{zanella2024mta,sheng2025rtpt,zhao2024rlcf} are not permutation-invariant functions of the
sorted probability vector, so \cref{app:matched2}'s symmetry argument does not apply to them and
we do not claim it does. They are nonetheless subject to the same empirical question, and the
image-feature statistics in \cref{tab:blind} are our proxy for it: on reducible errors they are
displaced from chance in the anti-predictive direction on both encoders. We did not run MTA
itself. Its published comparison is against confidence filtering on the full test population,
not on reducible errors, so it does not settle the question the audit asks, and we name it as
the first missing baseline rather than claim our conclusion covers it.

\paragraph{Delineation from ZERO.}
ZERO's Proposition~2.1 establishes that the argmax of $\pbar$ is invariant to the
marginal-entropy update. That is their result, not ours, and the corresponding row of
\cref{tab:resid} is a reproduction rather than a contribution. What we add is orthogonal: the
pairwise gate-times-log-ratio form of \cref{eq:thmA}, the three-case dissent criterion of
\cref{eq:thmB}, the stage-wise attribution separating selection from aggregation from the
update, and the audit of the selection signal that both ZERO and its successors retain.

\section{Broader impact}
\label{app:impact}

This work studies an existing test-time adaptation pipeline and reports where its reliability
signal fails. The practical recommendations, deleting a backward pass that contributes nothing
measurable and not treating prediction confidence as a per-view reliability signal, reduce
compute and discourage a form of unwarranted trust in model confidence, both of which we regard
as beneficial. The crop-area signal we propose uses no labels and no additional data, and
records only a geometric parameter the augmentation pipeline already samples, so it introduces
no new privacy surface. The main risk we can identify is misreading the audit: a score that is
near chance on reducible errors is not thereby useless overall, and \cref{app:failed} records
that we made exactly this error ourselves before the control was run. Deployments that gate
predictions on confidence should not conclude from this paper that confidence carries no
information, only that it should not be relied on to identify which view of an image is
trustworthy.

\end{document}